\documentclass[review]{elsarticle}

\usepackage[utf8]{inputenc} 
\usepackage[T1]{fontenc}    
\usepackage{hyperref}       
\usepackage{url}            
\usepackage{booktabs}       
\usepackage{amsfonts}       
\usepackage{nicefrac}       
\usepackage{microtype}      
\usepackage{amsmath}
\usepackage{amssymb}
\usepackage{mathtools}
\usepackage{amsthm}
\usepackage{microtype}
\usepackage{graphicx}
\usepackage{subfigure}
\usepackage{xcolor} 

\theoremstyle{plain}
\newtheorem{theorem}{Theorem}[section]

\newtheorem{lemma}[theorem]{Lemma}

\theoremstyle{definition}

\theoremstyle{remark}

\title{Conditional expectation with regularization for missing data imputation}

\usepackage{algorithm}
\usepackage{algcompatible}
\begin{document}
\begin{frontmatter}

 
\author[eighthaddress]{Anh Mai Vu$^{\dagger }$}
\author[eighthaddress]{Tu T. Do$^{\dagger }$}
\author[firstaddress]{Thu Nguyen$^{\dagger }$\footnote{$\dagger$ denotes equal contribution}}
\author[eighthaddress]{Nhan Thanh Phan}
\author[secondaddress]{Nitesh V. Chawla}
\author[firstaddress]{Pål Halvorsen}
\author[firstaddress]{Michael A. Riegler}
\author[eighthaddress,ninthaddress]{Binh T. Nguyen\corref{correspondingauthor}}
\cortext[correspondingauthor]{Corresponding author}
\ead{ngtbinh@hcmus.edu.vn}
\address[firstaddress]{Simula Metropolitan, Oslo, Norway} 
\address[secondaddress]{University of Notre Dame,
    Notre Dame, Indiana, USA}
\address[eighthaddress]{University of Science, Ho Chi Minh City, Vietnam}
\address[ninthaddress]{Vietnam National University, Ho Chi Minh City, Vietnam}  
\begin{abstract}
Missing data frequently occurs in datasets across various domains, such as medicine, sports, and finance.
Very often, the missing values are imputed using a method that yields a low root mean square error (RMSE) between the imputed and the true values, before subsequent analysis and inference. 
In addition, for some critical applications, it is also often a requirement that the imputation method is scalable, i.e., computationally feasible for large datasets, and the logic behind the imputation is explainable, which is especially difficult for complex methods that are, for example, based on deep learning. 
%
%
Based on these considerations, we propose a new algorithm named \textit{conditional Distribution-based Imputation of Missing Values with Regularization} (DIMV). DIMV operates by determining the conditional distribution of a feature that has missing entries, using the information from the fully observed features as a basis. 
As will be illustrated via experiments in the paper, DIMV (i) gives a low RMSE for the imputed values compared to state-of-the-art methods; (ii) is fast and scalable; (iii) is as explainable as coefficients in a regression model, allowing reliable and trustable analysis, makes it a suitable choice for critical domains where understanding is important such as in medical fields and finance; (iv) can provide an approximated confidence region for the missing values in a given sample; (v) is suitable for both small and large scale data; (vi) in many scenarios, does not require a huge number of parameters as deep learning approaches; (vii) can handle multicollinearity in imputation effectively;  and (viii) is robust to the normally distributed assumption that its theoretical grounds rely on.

\end{abstract}

\end{frontmatter}

 \section{Introduction}
 Missing data is a frequent problem in practice. For example, in a survey, some participants may not answer all the questions, making some entries in the dataset missing, or in medical records, some test results are missing for some patients. 
 There have been various methods trying to tackle the problem. Most of them are imputation methods that try to recreate and fill in the missing values~\cite{emmanuel2021survey}. However, in most works on missing data, the explainability of scalability of imputation has not been paid much attention to.
 

Meanwhile, the importance of explainable AI is increasingly recognized across various sectors.
Taking medical imaging is an example where physicians rely on explainability to evaluate and potentially improve diagnoses based on machine outputs.
However, a significant challenge arises when these explanation methods encounter missing data in the dataset, an aspect they often overlook.
The occurrence of missing data inherently introduces additional uncertainty into the model.
Depending on the imputation values used to fill in the missing data, explanations and results can vary, potentially leading to inaccurate evaluations of a patient's health status.
Such issues are not exclusive to healthcare but extend to other domains, including banking, biology, and more.

Thus, being able to explain the imputation of missing values in the data is an essential step toward more trustworthy data analysis.
This motivates us to introduce \textit{conditional Distribution based Imputation of Missing Values} (DIMV) algorithm, an algorithm that is able to find the conditional distribution of the features with missing values based on fully observed features in a fashion similar to the Expectation-Maximization algorithm. However, DIMV only conditions upon selected relevant features, and moreover, the mean and covariance matrix are estimated directly based on the DPER algorithm \cite{NGUYEN2022108082}, which gives DIMV an advantage in speed compared to EM. In addition, the imputation step of DIMV can be interpreted as Ridge regression, with coefficients giving direct explainability as regression coefficients. 
Furthermore, note that by the multivariate central limit theorem, the multivariate normal distribution is a good approximation to various distributions. Therefore, as will be illustrated in the experiments, our method is robust against this assumption.

In short, our contributions are as follows: (i) We introduce the DIMV algorithm for explainable missing data imputation that can achieve a low root mean square error (RMSE) between the imputed values and the actual values; (iii) We provide theoretical grounds for our approach; (iv) We show the relation between DIMV and EM algorithm, as well as Ridge regression and that DIMV is capable of handling multicollinear data; (v) We show that the imputation in DIMV is explainable in a regression coefficient like manner; (vi) We illustrate the prominent performance of the proposed approaches and the robustness of DIMV to the assumption of normal distribution via various experiments; and (vii) We analyze the advantages of the proposed methods and potential research directions.


\section{Related Work}\label{sec:related_work}
The most common way to deal with missing values is by using some imputation method to fill in the missing values. Through matrix decomposition or matrix completion, as in Polynomial Matrix Completion \cite{fan2020polynomial}, ALS \cite{hastie2015matrix}, and Nuclear Norm Minimization \cite{candes2009exact}, continuous data can be made complete and then can be analyzed with regular data analysis procedures. In addition, many methods are based on regression or clustering, such as the CBRL and CBRC \cite{m2020cbrl}, which rely on Bayesian Ridge Regression, 
and cluster-based local least square method \cite{keerin2013improvement}. 
When dealing with large data, some deep learning imputation techniques have gained attention due to good performance \cite{choudhury2019imputation, garg2018dl, mohan2021graphical}. Note that different imputation approaches may fill in different values for each missing entry. Therefore, sometimes, modeling the uncertainty for each missing entry is also of interest. In such cases, Bayesian or multiple imputation techniques, such as Bayesian principal component analysis-based imputation \cite{audigier2016multiple} and multiple imputations using Deep Denoising Autoencoders \cite{gondara2017multiple}, are preferred. 
Moreover, some classes of tree-based techniques can naturally handle missing data by prediction, such as missForest \cite{stekhoven2012missforest}, the DMI algorithm \cite{rahman2013missing}, 
DIFC \cite{nikfalazar2020missing}. In addition, some recent methods that can handle mixed data are SICE \cite{khan2020sice},  FEMI \cite{rahman2016missing}, and HCMM-LD \cite{murray2016multiple}.

In recent years, some studies tend to combine the imputation and the target task into one problem or adapt the model to learn directly from missing data. For example, Dinh et al.~\cite{dinh2021clustering} integrate the imputation and clustering steps into a single process consisting of three phases: initialization, imputation, and clustering. 
Another example is the work of  \cite{ghazi2018robust}, which models the course of diseases using an LSTM architecture that includes built-in handling of missing data in order to account for the missing data in the input and the targets. This is achieved by using batch gradient descent with back-propagation through a time algorithm. Li et al.~\cite{LI2020304} suggested a technique for the bi-clustering problem that can handle missing data (the problem of partitioning rows and columns of a rectangular data array into homogeneous subsets simultaneously). 
Learning directly from the data may have advantages in speed or help reduce the storage costs of storing one model for imputation and another for the target task. However, the complexity of these approaches and the fact that they are not readily available in packages and might not generalize well across different datasets hinder their applications \cite{nguyen4260235pmf}. On the other hand, imputation makes the data complete and easier to generalize across different datasets and can be analyzed with various data analysis techniques. 

However, not all of the mentioned approaches are explainable, and to our knowledge, only very few works have been paid to the explainability of missing data imputation. For example, Hans et al.~\cite{hans2023explainable} present an explainable imputation method based on the association constraints in data. Here, the explanations for imputations come directly from the constraints used. Compared to DIMV, such a constraint-based method requires known relationships or restrictions, which may not always be available, limiting their applicability. In addition, the paper conducts experiments on only small datasets. Hence, the scalability of this method needed to be investigated.
The DIMV imputation method relies on parameter estimation to find the conditional distribution of a missing entry based on the observed ones. 
For this, there have been some works on direct parameter estimation instead of using optimization approaches such as the EM algorithm or imputing missing values and then estimating the parameters. Specifically, Nguyen et al.~\cite{NGUYEN20211} introduced the EPEM algorithm to estimate the MLEs for multiple class monotone missing data when the covariance matrices of all the classes are assumed to be equal. Further, DPER \cite{NGUYEN2022108082} is meant for a more general case, where missing data can occur in any feature by considering pairs of features to estimate the entries in the mean and covariance matrices. 
Besides the advantage of computational speed and estimation accuracy, such direct approaches also allow for deriving the distribution of the estimates under the normally distributed assumption of the data. 


\section{Preliminaries: DPER algorithm for parameter estimation}
In this section, we present the \textbf{DPER algorithm} for the estimation of the mean and covariance matrix for a dataset that \textbf{\textit{consists of a single class}}. The algorithm is based on Theorems 4.1 and 4.2 by Nguyen et al.~\cite{NGUYEN2022108082}.

\begin{theorem}\label{bivariatemle_single}
	Assume that we have a set of i.i.d observations from a bivariate normal distribution with mean  $\boldsymbol {\mu } = (\mu_1, \mu_2)^T$ and covariance matrix 
		$\boldsymbol {\Sigma } =  \begin{pmatrix}
			\sigma_{11}&\sigma_{12}\\
			\sigma _{21} & \sigma _{22}
		\end{pmatrix}.$
	Arrange the data into the following pattern
	\begin{equation}\label{blform}
	\small
		\boldsymbol{x} = \begin{pmatrix}
			{x}_{11} & ... & {x}_{1m}&{x}_{1m+1}&...&{x}_{1n}&*&...&*\\
			{x}_{21}&...&{x}_{2m}& *& ...& *& {x}_{2n+1}&...&{x}_{2l}
		\end{pmatrix}.
	\end{equation}
	So, each column represents an observation, and $x_{ij} \in \mathbb{R}$ is an entry, i.e., each observation has two features, there are $m$ samples that have both features observed, $n$ samples that have the first feature observed, and $m+l-n$ samples that have the second feature observed. 
	
	Let $L$ be the likelihood of the data and	
	\begin{align}
		s_{11} &= \sum_{j=1}^{m}(x_{1j}-\hat{\mu}_1)^2,\;\\
		s_{12} &= \sum_{j=1}^{m}(x_{2j}-\hat{\mu}_2)(x_{1j}-\hat{\mu}_1),\;\\
		s_{22} &= \sum_{j=1}^{m}(x_{2j}-\hat{\mu}_2)^2.
	\end{align}
	
	Then, the resulting estimators obtained by maximizing $L$ w.r.t $\mu_1, \sigma_{11},\mu_2, \sigma_{22}$, and $\sigma_{12}$ are:
	\begin{align*}
		\hat{\mu}_1 & = \frac{1}{n}\sum_ {j=1}^{n}x_{1j},\;\;
		\hat{\mu}_2 = \frac{\sum_ {j=1}^{m}x_{2j}+ \sum_ {j=n+1}^{l}x_{2j}}{m+l-n},	 \\
		\hat{\sigma}_{11} & = \frac{\sum_ {j=1}^n(x_{1j}-\hat{\mu}_1)^2}{n},
		\hat{\sigma}_{22}  = \frac{\sum_ {j=1}^{m}(x_{2j}-\hat{\mu}_2)^2+ \sum_ {j=n+1}^{l}(x_{2j}-\hat{\mu}_2)^2}{m+l-n},	
	\end{align*}

	and $\hat{\sigma} _{12}$, where $\hat{\sigma} _{12}$ is a solution of
		\begin{align}
    	s_{12} \sigma_{11} \sigma_{22} + {\left( m\sigma_{11} \sigma_{22}  -  s_{22}\sigma_{11} - s_{11}\sigma_{22} \right)}\sigma_{12}  
     + s_{12}\sigma_{12}^2- m\sigma_{12}^3, \label{equation:3rd poly} 
        \end{align}
that maximizes	
	\begin{align}\label{etacov-sing}
		\eta = C-\frac{1}{2}m\log \left(\sigma_{22}-\frac{\sigma_{12}^2}{\sigma_{11}}\right)-\frac{1}{2} \left( s_{22} - 2\frac{\sigma_{12}}{\sigma_{11}} s_{12}+\frac{\sigma_{12}^2}{\sigma _{11}^2}s_{11}\right)\left(\sigma_{22}-\frac{\sigma_{12}^2}{\sigma_{11}}\right)^{-1}.
	\end{align}
\end{theorem}
\begin{theorem}\label{unique-single}
		Solving
		\begin{equation}\label{der01}
			\frac{d\eta}{d\sigma_{12}} = 0
		\end{equation}
		can be reduced to solving the following equation 
		\begin{align}
    	\small
    	s_{12} \sigma_{11} \sigma_{22} + {\left( m\sigma_{11} \sigma_{22}  -  s_{22}\sigma_{11} - s_{11}\sigma_{22} \right)}\sigma_{12} 
     + s_{12}\sigma_{12}^2- m\sigma_{12}^3 \label{equation:3rd poly} 
        \end{align}
		which has at least one real root. In addition, the global maximum is a real solution to that equation, provided that
\begin{equation} 
	-\frac{s_{22}\sigma_{11}+s_{11}\sigma_{22}}{2\sqrt{\sigma_{11}\sigma_{22}}} \neq
	s_{12} \neq \frac{s_{22}\sigma_{11}+s_{11}\sigma_{22}}{2\sqrt{\sigma_{11}\sigma_{22}}}.
\end{equation}
\end{theorem}
Based on the above theoretical ground, the DPER algorithm for single-class randomly missing data is presented in Algorithm \ref{alg:dimv1f}, where the estimation of the entries in the covariance matrix $\mathbf{\Sigma}$ is conducted in pairs of features. Since the input to DIMV is centered, we simplify the equations in Theorem 1.1 using the fact that for centered data, $\mu_1=\mu_2=0$.

\begin{algorithm}[tb]
    \caption{DPER algorithm for single-class randomly missing data}
    \label{alg:dimv1f}
    {\bfseries Input:} A data set of $p$ features.\\
    {\bfseries Output:} $\hat{\boldsymbol {\mu}}$ and $\hat{\boldsymbol {\Sigma}}$.\\
    {\bfseries Procedure:}
    
    \begin{algorithmic}[1]        
        \STATE Estimate $\boldsymbol{\mu}$: $\hat{\mu}_i$ is the sample mean of all the available entries in the $i^{th}$ feature. 
        \FOR {$1\le i\le p$}
            \STATE Compute $\hat{\sigma}_{ii}$: the uncorrected sample variance of all the available entries in the $i^{th}$ feature. 
            \FOR {$1\le j < i$}
                \STATE Compute $\hat{\sigma}_{ij}$ based on Equation  (\ref{der01}).
                \IF {there are two solutions maximizing the function}
                    \STATE Choose the one closest to the estimate based on the case deletion method.
                \ENDIF
            \ENDFOR
        \ENDFOR
        \STATE $\hat{\boldsymbol {\Sigma}}=(\sigma_{ij})_{i,j=1}^p$
        \STATE \textbf{return} $\hat{\boldsymbol {\mu}}, \hat{\boldsymbol {\Sigma}}$.
    \end{algorithmic}
\end{algorithm}

	
	
	
		
		
		
		
		
		

Recall that in the \textit{case deletion method}, the sample that has one or more missing values is deleted. In this setting, during the estimation of $\hat{\sigma}_{ij}$ using case deletion, we delete any $(i,j)^{th}$ pair that has at least one missing entry. 

\section{Motivation and theoretical grounds}\label{sec-motivation}
In this section, we present the motivation and theoretical grounds for the DIMV algorithm, which will be presented in the next section. The basic idea of DIMV is to use the Gaussian conditional formula, along with feature selection and $L_2$ norm regularization, to make imputation explainable, scalable, and robust to multicollinearity. The Gaussian conditional formula is as follows \cite{johnson2002applied}:
\begin{theorem}\label{theo:condi}
 Suppose $\mathbf{y} \sim N_p(\boldsymbol {\gamma }, \boldsymbol {\Sigma })$. Let $\mathbf  {y}=(\mathbf  {y}_{1}^T,{\mathbf  {y}}_{2}^T)$, where $\mathbf{y}_1 \in \mathbb{R}^q$ and $\mathbf{y}_2 \in \mathbb{R}^{(p-q)}$.
Partition accordingly
\begin{equation*}
{\boldsymbol {\gamma }}={\begin{bmatrix}{\boldsymbol {\gamma }}_{1}\\{\boldsymbol {\gamma }}_{2}\end{bmatrix}}, 
\boldsymbol\Sigma
=
\begin{bmatrix}
\boldsymbol\Sigma_{11} & \boldsymbol\Sigma_{12} \\
\boldsymbol\Sigma_{21} & \boldsymbol\Sigma_{22}
\end{bmatrix}
\end{equation*}
Then, the distribution of $\mathbf{y}_1$ conditional on $\mathbf{y}_2=a$ follows multivariate normal distribution with mean
$\bar{\boldsymbol\gamma}
=
\boldsymbol\gamma_1 + \boldsymbol\Sigma_{12} \boldsymbol\Sigma_{22}^{-1}
\left(
\mathbf{a} - \boldsymbol\gamma_2
\right)$
and covariance matrix
${\overline {\boldsymbol {\Sigma }}}={\boldsymbol {\Sigma }}_{11}-{\boldsymbol {\Sigma }}_{12}{\boldsymbol {\Sigma }}_{22}^{-1}{\boldsymbol {\Sigma }}_{21}.$
\end{theorem}
Suppose that we have a vector $\mathbf{U}\in \mathbb{R}^p$ that follows multivariate normal distribution with mean $\boldsymbol {0} $ and covariance matrix $\mathbf{\Sigma}$, partition 
\begin{equation}\label{eqx}
    \mathbf{U} = \begin{pmatrix}
    \mathbf{U}_o\\
    \mathbf{U}_m
    \end{pmatrix}
\end{equation}
where $\mathbf{U}_o \in \mathbb{R} ^q$ represents the \textit{observed} part and $\mathbf{U}_m \in \mathbb{R} ^{p-q}$  represents the \textit{missing} part of the vector. Next, we partition $\mathbf{\Sigma}$ accordingly
\begin{equation}
    \mathbf{\Sigma}=\begin{pmatrix}
    \mathbf{\Sigma}_o & \mathbf{\Sigma}_{om}\\
    \mathbf{\Sigma}_{om}^T & \mathbf{\Sigma}_m
    \end{pmatrix}
\end{equation}
Then, as a result of the theorem, the conditional distribution of $\mathbf{U}_m$ given that $\mathbf{U}_0 = \mathbf{u}_o$ is normal with mean $\boldsymbol {\mu }_{m|o}= \mathbf{\Sigma}_{om}^T\mathbf{\Sigma}_o^{-1} \mathbf{u}_o$ and covariance matrix. 
    $\mathbf{\Sigma}_{m|o} =  \mathbf{\Sigma}_m - \mathbf{\Sigma}_{om}^T\mathbf{\Sigma}_o^{-1}\mathbf{\Sigma}_{om}$.
Therefore, we can estimate the missing part $\mathbf{U}_m$ of $\mathbf{U}$ with the conditional mean $\boldsymbol {\mu }_{m|o}$, i.e. $\hat{\mathbf{U}}_m = \boldsymbol {\mu }_{m|o}$, and the variation is reflected through the covariance matrix $\mathbf{\Sigma}_{m|o}$.

\paragraph{Prediction in blocks of samples.} If the data can be partitioned in blocks, then computation in the matrix form can help significantly speed up the computation compared to imputing sample by sample. So, for the computation in blocks, note that if there exists another sample $\mathbf{u}'^T=(\mathbf{u}_o'^T, \mathbf{u}_m'^T)$ that has the same missing pattern as $\mathbf{u}$ then by the same imputation strategy, $\hat{\mathbf{u}}_m' = \mathbf{\Sigma}_{om}^T\mathbf{\Sigma}_o^{-1} \mathbf{u}_o'$, and the conditional covariance matrix is also $ \mathbf{\Sigma}_m - \mathbf{\Sigma}_{om}^T\mathbf{\Sigma}_o^{-1}\mathbf{\Sigma}_{om}$. Hence, the prediction can be done in blocks. Suppose we have a block $\boldsymbol{Z}$, where each column is a sample with the same missing pattern as $\mathbf{x}$. Partition
    $\boldsymbol{Z} = \begin{pmatrix}
    \boldsymbol{Z}_o\\
    \boldsymbol{Z}_m
    \end{pmatrix}.$
Then, $\boldsymbol{Z}_m$ can be estimated by
$\mathbf{\Sigma}_{om}^T\mathbf{\Sigma}_o^{-1} \mathbf{Z}_o,$
and the covariance matrix is $\mathbf{\Sigma}_{m|o}$.


    

\textbf{The effects of conditioning on redundant features.}
In practice, however, while trying to predict a missing entry in a feature $f$, there can be many features in the dataset that are not related to that feature. Prediction of the missing entries in $f$ based on these unrelated features may induce computational error in inverse matrix computation, noises, and computational cost for larger matrix inversion. These issues can be ameliorated by \textbf{\textit{not}} conditioning on the features that have low correlations with $f$. 
The following theorem illustrates the effect of conditioning on features with low correlation, 
\begin{theorem}\label{theorem-unrelated}
    Assume we have an observation $\mathbf{x}$ that comes from a distribution with zero mean and covariance matrix $\mathbf{\Sigma}$. Suppose that we can partition $\mathbf{x}, \mathbf{\Sigma}$ into
    \begin{equation}
        \mathbf{x}=\begin{pmatrix}
            \mathbf{x}_o\\
            \mathbf{x}_\epsilon\\
            \mathbf{x}_m
        \end{pmatrix}, \;
        \mathbf{\Sigma}=\begin{pmatrix}
            \mathbf{\Sigma}_{o} & \mathbf{\Sigma}_{o\epsilon} & \mathbf{\Sigma}_{om}\\
            \mathbf{\Sigma}_{\epsilon o} & \sigma_\epsilon & \epsilon \\
            \mathbf{\Sigma}_{mo} & \epsilon &\sigma_m
        \end{pmatrix}        
    \end{equation}
    where $\mathbf{x}_m \in \mathbb{R}$ is a feature that contains missing values, $\mathbf{x}_o$ corresponding to the observed partition that is highly correlated with the missing entry $\mathbf{x} \in \mathbb{R}$, and $\mathbf{x}_\epsilon \in \mathbb{R}$ is observed but has a low correlation of $\epsilon$ with $\mathbf{x}_m$. In addition, $\mathbf{\Sigma}_o$ is the partition of $\mathbf{\Sigma}$ that corresponds to the observed features, $\sigma_\epsilon, \sigma_m$ is the variance of $\mathbf{x}_\epsilon, \mathbf{x}_m$, respectively. Also, $\mathbf{\Sigma}_{\epsilon o}, \mathbf{\Sigma}_{mo}$ is the covariance matrix between $\mathbf{x}_\epsilon$ and $\mathbf{x}_o$, $\mathbf{x}_m$ and $\mathbf{x}_o$, respectively.

    Let $\hat{\mathbf{x}}_m = \hat{\mathbf{\Sigma}}_{om}^T \hat{\mathbf{\Sigma}}_o^{-1} \mathbf{x}_o $ be the imputed value based on conditioning on $\mathbf{x}_o$. In addition, let $\hat{\mathbf{x}}_{\epsilon}$ be the imputed value of $\mathbf{x}_m$ based on conditioning on both $\mathbf{x}_o$ and $\mathbf{x}_\epsilon$. Moreover, suppose that $\sigma_\epsilon\neq \hat{\mathbf{\Sigma}}_{\epsilon o} \hat{\mathbf{\Sigma}}_o^{-1}\hat{\mathbf{\Sigma}}_{o\epsilon}.$
    Then, the difference between the two imputed values is
    \begin{equation}
        \hat{\mathbf{x}}_\epsilon-\hat{\mathbf{x}}_m = \frac{(\hat{\mathbf{\Sigma}}_{mo} \hat{\mathbf{\Sigma}}_o^{-1} \hat{\mathbf{\Sigma}}_{o\epsilon} -\epsilon) (\hat{\mathbf{\Sigma}}_{\epsilon o} \hat{\mathbf{\Sigma}}_o^{-1} \mathbf{x}_o-\mathbf{x}_\epsilon) }{\sigma_\epsilon -  \hat{\mathbf{\Sigma}}_{\epsilon o} \hat{\mathbf{\Sigma}}_o^{-1}\hat{\mathbf{\Sigma}}_{o\epsilon} }.
    \end{equation}
\end{theorem}

It is worth noting that we use the subscript $_o$ instead of $_{o_\mathcal{F}}$ in the above statement to simplify the notations. The proof of this statement is available in  \ref{proof-unrelated}.
The theorem shows that the extent to which a redundant feature can affect an imputed value also depends on the correlations between the redundant feature and the observed features.

\paragraph{Conditioning with regularization.}
Recall that for an ordinary least square problem\textbf{\textit{ without missing values}}
   $y = \mathbf{X} \beta +\eta,$ where $\mathbf{X}$ is the \textit{\textbf{design matrix}} (i.e., each row represents a sample and each column represents a feature) and $\eta \sim \mathcal{ N}(0, \sigma^2I)$ is the noise,
we have 
    $Cov(\mathbf{X},y) = Cov(\mathbf{X},\mathbf{X}\beta+\eta) = Cov(\mathbf{X}) \beta.$
Hence, it is well known that
    $\beta = Cov(\mathbf{X})^{-1} Cov(\mathbf{X},y)$.

Next, note that if both $\mathbf{X}$ and $\mathbf{y}$ are centered then $Cov(\mathbf{X}) = \mathbf{X}^T\mathbf{X}$. Therefore, for Ridge regression without missing data, 
    $\hat{\beta} = (\mathbf{X}^T\mathbf{X}+\alpha  I)^{-1}\mathbf{X}^Ty = (Cov(\mathbf{X})+\alpha I)^{-1}Cov(\mathbf{X},y)$ (Here, $\alpha$ is the regularization parameter, and $I \in \mathbb{R}^{d}$ is the identity matrix).
Hence, to avoid overfitting and improve the generalizability when predicting missing values, we can add the regularization term $\alpha$ for the prediction of a sample and imputation in a block as follows
\begin{equation}\label{equal:con_reg}
  \mathbf{\Sigma}_{om}^T(\mathbf{\Sigma}_o +\alpha  I )^{-1} \mathbf{u}_o, \;\; \boldsymbol{Z}_m = 
\mathbf{\Sigma}_{om}^T(\mathbf{\Sigma}_o +\alpha  I ) ^{-1} \mathbf{Z}_o.
\end{equation}

The above analysis leads to the DIMV algorithm described in the next section.
\section{DIMV algorithm}\label{sec-dimv-alg}





\begin{algorithm}[!h]
    \caption{FeatureSelection}
    \label{algo:FS}
    {\bfseries Input:} $\Sigma$, $f$, $s \in \mathbb{R}^{(1\times q)}$, 
    set of feature selection parameters $\mathbf{P}$ including:
    \begin{itemize}
        \item $\tau \in [0, 1):$  correlation threshold, 
        \item $k$ a predefined number of top features selected for the edge case  (when there are no features that meet the correlation threshold condition).  
    \end{itemize} 
    {\bfseries Procedure:}
    \begin{algorithmic}[1]
        \STATE $\mathcal{F}_{\tau} \leftarrow$ set of features in $q\setminus f$ whose absolute value of Pearson's correlation coefficient with $f$ is greater than $\tau$ and $\forall j \in \mathcal{F}_{\tau}, s_{j}\neq $ NA  (NA: missing values); $\triangleright$ $s_{j} \leftarrow$ value in $s$ correspond to feature $j$;  
        
        \IF {$s_{F\tau}$ is empty} 
            \STATE $\mathcal{G} \leftarrow $ set of features in $q \setminus \mathcal{F} $ and $\forall g \in G, s_{g}\neq$ NA  (NA: missing values), 
            \STATE $\mathcal{F} \leftarrow \mathcal{G}$,
        \ELSE
            \STATE $\mathcal{F} \leftarrow \mathcal{F}_{\tau}$, 
        \ENDIF 
        \STATE \textbf{return} $\mathcal{F}$.
            
    \end{algorithmic}
\end{algorithm}
The \textbf{DIMV} algorithm, presented in Algorithm \ref{algo:dimv}, requires the training and testing data to be centered. Note that centering and scaling are commonly used techniques for preprocessing, and the means and covariance matrix of the original data can be inverted easily from the scaled version. Therefore, even though the input to DIMV should be centered, the data can be easily inverted back to the original, not centered version using the fact that $\mathbb{E}(\mathbf{Z}+a)=\mathbb{E}(\mathbf{Z})+a$ for any random variable $\mathbf{Z}$ and any constant $a$.  

In addition, DIMV also requires an indication of the initialization value,  $init\_with\_zero$; when set to True, missing values in the test set will be initialized as 0; otherwise, they will be kept as missing values. This parameter helps to speed up the block imputation process for complex missing patterns.  When it is set to True,  all the remaining samples with missing data in feature $f$ are initialized with 0 and stacked into $Z$. 
This reduces the computational cost of stacking the data. Specifically, when the data has randomly missing patterns, the number of missing patterns at each sample can be as high as $2^{q}$, which means, in the worst case, it will cost $n$ operations for one feature. 
Therefore, when dealing with large datasets with random missing patterns,
it is recommended to set the initialization value to $0$ to speed up the process. The algorithm also requires a set of parameters, $\mathbf{P}$, for the feature selection process.

DIMV starts by estimating the covariance matrix $\mathbf{\Sigma}$ using DPER algorithm  \cite{NGUYEN2022108082} from the training set (step 1). Next, the algorithm proceeds to impute each feature in the test set by iterating through the features with missing values. The imputation for each feature $f$ is performed in blocks of samples, where each block is defined for a particular feature by stacking samples with the same missing patterns, starting with an example $s$, and searching for similar features with the same missing pattern for imputation. The imputation for each feature $f$ is carried out based on Theorem \ref{theo:condi}, as follows:
\begin{enumerate}
    \item First, the algorithm collects a set of features $\mathcal{F}$ using the \textbf{FeatureSelection} algorithm presented in algorithm \ref{algo:FS}. The algorithm searches for features that are observed in sample $s$ and have a correlation with $f$ larger than a threshold $\tau \in [0, 1)$, among the remaining features. If no feature meet the requirement, the top $k$ features with the highest correlation with $f$ are chosen. A value of $\tau$ = 0 indicates that no feature selection is carried out, and all remaining observed features at $s$ are chosen.  
    \item After having $\mathcal{F}$, DIMV starts to impute a block of the sample (lines 12 - 15). Here, by the samples that have the same missing patterns, which form a block, we mean the samples in the same block should satisfy two requirements. Let $\mathcal{M}_{s}$, $\mathcal{M}_{i}$ denote the missing features in sample $s$ and $i$, respectively. Then, the first requirement is that $\mathcal{M}_{i}\subseteq \mathcal{M}_{s}$, this ensures that there are no potential features in $\mathcal{M}_{s}$ missed when for imputing for $i$. Let $\mathcal{A}_{i}$ denote the available features in sample $i$. The second requirement is that $\mathcal{F}\subseteq \mathcal{A}_{i}$, which means $\mathcal{F}$ is also observed features at $i$. After imputing this block, we remove it from the $\mathcal{D}_{f}$ (line 16). 

\item  Finally, the algorithm uses conditional expectation regularization to compute the estimations of $\boldsymbol{Z}_{f}$ as specified in equation \ref{equal:con_reg} (line 13 to 15), $\boldsymbol{Z}_{o_\mathcal{F}}$, and two submatrices $\hat{\mathbf{\Sigma}}_{o\mathcal{F}}$ and $\hat{\mathbf{\Sigma}}_{o\mathcal{F}f}$, which can be filtered from the covariance matrix $\hat{\mathbf{\Sigma}}$. Here, $\boldsymbol{Z}[, \mathcal{F}]$ consists of columns for features $\mathcal{F}$ in matrix $\boldsymbol{Z}$ and $\hat{\mathbf{\Sigma}}[f, \mathcal{F}]$ slices matrix $\hat{\mathbf{\Sigma}}$ at row $f$, keeping only the columns in set $\mathcal{F}$. The regularization term $\alpha$ can be selected by a grid search. For each alpha in the grid, the RMSE is computed between the conditional expectation for observed positions on the train set and the original matrix. The $\alpha$ value yields the smallest RMSE is chosen as the optimal one. To tune this parameter, a subset of the train set can be utilized instead of the entire dataset. 
\end{enumerate} 

\begin{algorithm}[]
    \caption{DIMV algorithm}
    \label{algo:dimv}
    
    \textbf{Input:} 
    \begin{itemize}
        \item \textbf{\textit{centered}} training samples $\mathbf{X}_{train}$, 
        \item \textbf{\textit{centered}} test samples $\mathbf{X}_{test} \in \mathbb{R}^{n \times q}$, 
        \item $init\_with\_zero$ (True/False): if set to True, missing values in the test set will be initialized as 0,
        \item set of feature selection parameters $\mathbf{P}$,
        \item $\alpha$: $L_2$ regularization parameter (to be fine-tuned).
    \end{itemize}
    
    \textbf{Procedure:}
    \begin{algorithmic}[1]
        \STATE $\hat{\mu}, \hat{\mathbf{\Sigma}} \gets \text{DPER}(\mathbf{X}_{train})$
        \FOR {index $f$ of the feature that belongs to the set of features with missing values in $\mathbf{X}_{test}$}
            \IF {$init\_with\_zero$ = True}
                \STATE $\mathbf{X}_{init} \leftarrow $  input matrix $\mathbf{X}_{test}$ with missing values initialized by 0,
                \STATE $\mathcal{D}_{f} = \mathbf{X}_{init}$,
            \ELSE 
                \STATE $\mathbf{X}_{init} = \mathbf{X}_{test}$,
                \STATE $\mathcal{D}_{f} \leftarrow $  set of samples that have missing values in feature $f$,
            \ENDIF 
            \FOR {sample $s \in D_f$; $s \in \mathbb{R}^{1 \times q}$}   \Comment{for loop only have 1 iteration if $init\_with\_zero$ = True}
            
                \STATE $\mathcal{F} \leftarrow $ \textbf{FeatureSelection}($\hat{\Sigma}, f, s$ , $\mathbf{P}$); $\mathcal{F} \supseteq (q \setminus f)$,
             
                \STATE $\boldsymbol{Z} \leftarrow $  stack of samples in $D_f$ that have the same missing pattern  
                
                \STATE $\boldsymbol{Z}_{o_{\mathcal{F}}} = \boldsymbol{Z}[, \mathcal{F}]$ and $\boldsymbol{Z}_{f} = \boldsymbol{Z}[, f]$ are the corresponding submatrices of $\boldsymbol{Z}$,

                \STATE $\hat{\mathbf{\Sigma}}_{o\mathcal{F}f} = \hat{\mathbf{\Sigma}}[f, \mathcal{F}]$ and $\hat{\mathbf{\Sigma}}_{o\mathcal{F}} = \hat{\mathbf{\Sigma}}[\mathcal{F}, \mathcal{F}]$ are the corresponding submatrices of $\hat{\mathbf{\Sigma}}$, 
                
                \STATE Impute missing values in $\boldsymbol{Z}_f$ by $\hat{\boldsymbol{Z}}_f = \hat{\boldsymbol {\mu}}_{f|o_\mathcal{F}} = \hat{\mathbf{\Sigma}}_{o_\mathcal{F}f}^T (\hat{\mathbf{\Sigma}}_{o_\mathcal{F}} + \alpha I)^{-1}\boldsymbol{Z}_{o_\mathcal{F}}$,
                
                \STATE $D_f = D_f \setminus \boldsymbol{Z}$, 
            \ENDFOR
        \ENDFOR
        \STATE \textbf{return} imputed $\mathbf{X}_{test}$.
    \end{algorithmic}
\end{algorithm}

\section{Properties and relation to previous works}
This section will discuss some properties of DIMV, especially the explainability and confidence region (CR) for the proposed DIMV approach and the relations of DIMV to previously known techniques. We discuss what happens when conditioning on all features (which means without any feature selection for each feature). The explainability and CR for DIMV are similar. 

\subsection{Relation to Ridge regression} 
From equation \ref{equal:con_reg}, we can see that the imputation step of DIMV can be interpreted as Ridge regression. 
However, this interpretation is in connection to Ridge regression without missing data. Meanwhile, it is essential to keep in mind that the overall DIMV procedure uses $\mathbf{\Sigma}$, which is estimated using DPER, and the overall DIMV procedure is different from Ridge regression due to the iterative nature of DIMV across the imputation of all features with missing values.

Therefore, the properties from Ridge regression, such as the expectation and covariance matrix of the coefficients, cannot be translated directly into the properties of the regression coefficient in DIMV. However, the imputation step in DIMV is also a shrinkage estimator. Hence, similar to Ridge regression, it reduces the variance of the estimate by introducing some bias (i.e., shrinks the estimates towards 0).

\subsection{Explainability} 
\paragraph{Explainability} For explainability, recall from section \ref{sec-motivation} that the conditional distribution of $\mathbf{U}_m$ given that $\mathbf{U}_0 = \mathbf{u}_o$ is normal with mean $$\boldsymbol {\mu }_{m|o}= 
\mathbf{\Sigma}_{om}^T(\mathbf{\Sigma}_o +\alpha I)^{-1} \mathbf{u}_o.$$ 
Moreover, DIMV estimates the missing part $\mathbf{X}_m$ of $\mathbf{X}$ with the conditional mean $\boldsymbol {\mu }_{m|o}$. As mentioned in the previous paragraph, the imputation step in DIMV can be interpreted as Ridge regression with a regression coefficient
$$\hat{\beta} = (\mathbf{\Sigma}_o +\alpha  I ) ^{-1}\mathbf{\Sigma}_{om}^T.$$
Therefore, the values of the coefficients determine the contribution of each observed feature to the imputation. In addition, these coefficients also indicate the strength and direction of the relationships between the observed and missing features.


\subsection{Confidence region (CR)} 
\textit{If $\alpha =0$, i.e., \textbf{when there is no regularization}}, the  CR for the missing entries can be derived based on Lemma \ref{d:chi_square} \cite{johnson2002applied}:
\begin{lemma}\label{d:chi_square}
    Let $\mathbf{V}$ be distributed as $N_p(\boldsymbol {0}, \mathbf {\Sigma })$ with $\left|\mathbf{\Sigma}\right|>0$. Then
        $$\left(\mathbf{V} - \boldsymbol{\mu}\right)^T\mathbf{\Sigma}^{-1}\left(\mathbf{V} - \boldsymbol{\mu}\right)\sim \boldsymbol{\chi}_p^2,$$ where $\boldsymbol{\chi}_p^2$ denotes the chi-square distribution with $p$ degrees of freedom. In addition, 
        the $N_p(\boldsymbol {0}, \mathbf {\Sigma })$ distribution assigns probability $1-\boldsymbol{\alpha}$ to the solid ellipsoid $\left\{\mathbf{v}: \left(\mathbf{v} - \boldsymbol{\mu}\right)^T\mathbf{\Sigma}^{-1}\left(\mathbf{v} - \boldsymbol{\mu}\right)\right\}$, where $\boldsymbol{\chi}_p^2\left(\boldsymbol{\alpha}\right)$ denotes the upper $(100 \boldsymbol{\alpha})$th percentile of the $\boldsymbol{\chi}_p^2$ distribution.
    \end{lemma}
Given a vector $\mathbf{U}$ that can be partitioned as in equation \ref{eqx}, then recall that $\mathbf{U}_o \in \mathbb{R}^q, \mathbf{U}\in \mathbb{R}^p$. Therefore,  $(\mathbf{U}_m|\mathbf{U}_0 = \mathbf{u}_o) \sim N_{p-q}(\boldsymbol {\mu }_{m|o}, \mathbf {\Sigma }_{m|o})$. By Lemma \ref{d:chi_square}, 
    \begin{equation}
(\mathbf{U}_m - \boldsymbol{\mu}_{m|o})^T\mathbf{\Sigma}_{m|o}^{-1}(\mathbf{U}_m - \boldsymbol{\mu}_{m|o}) \sim \boldsymbol{\chi}_{p-q}^2. 
    \end{equation}
    Hence, a $(1-\boldsymbol{\alpha})\%$ confidence region is
    \begin{equation*}
        \left[\mathbf{u}:(\mathbf{u} - \boldsymbol{\mu}_{m|o})^T\mathbf{\Sigma}_{m|o}^{-1}\left(\mathbf{u} - \boldsymbol{\mu}_{m|o}\right) \le \boldsymbol{\chi}_{p-q}^2(\alpha)\right].
    \end{equation*}
However, note that the derived CR is based on the assumption of normality. This means that for small sample datasets if the underlying data deviates significantly from normality, the confidence region may not be accurate. However, thanks to the Central Limit Theorem, the above CR can be a good approximation when the sample size is large enough.   


\subsection{Relation to EM algorithm for Gaussian mixture model (GMM)}
The EM algorithm for GMM consists of an Expectation step (E-step) that creates a function for the expectation of the log-likelihood evaluated using the current estimate for the parameters and a Maximization step (M-step) that finds parameters that maximize the expected log-likelihood found on the E-step. The M-step also uses the conditional Gaussian formula as in DIMV. However, one can note that DIMV only uses one Gaussian distribution instead of a mixture of Gaussians. Also, the estimate for the mean and covariance matrix is conducted by using the DPER algorithm for one class \cite{NGUYEN2022108082}. We reckon this helps reduce the computational cost compared to using a mixture of Gaussian, which is known to be computationally expensive \cite{delalleau2012efficient}. 

To summarize, DIMV can be considered as a combination of the ideas from EM, feature selection, and Ridge regression. DIMV shares some similarities with EM in the sense that both use the Gaussian conditional formula. However, DIMV uses only one Gaussian, and there is no loop for finding the mean and covariance matrix as in the EM algorithm. In addition, DIMV conducts feature selection for identifying relevant features to condition upon, and Ridge regression helps deal with overfitting and multicollinearity. 
\section{Experiments}\label{sec-experiment}
\subsection{Experiment settings}
\begin{table}[ht]
    \caption{Description of datasets used in the experiment}  
    \label{tab-datasets}
    \vskip 0.15in
    \begin{center}
    \begin{small}
    \begin{sc}
    \begin{tabular}{|c|c|c|c|c|c|}
        \hline
        Dataset  & \#Features & \#Sample & \#Missing type\\
        \hline 
        MNIST              &  784 & 70000 & Monotone Missing \\ 
        Fashion-MNIST      &  784 & 70000 & Monotone Missing  \\ 
        Yeast    &  8  & 1484 & Randomly Missing  \\
        Thyroid  &  5  & 215  & Randomly Missing  \\ 
        Seeds    &  7  & 210 & Randomly Missing   \\ 
        Iris     &  4  & 150  & Randomly Missing \\ \hline    \end{tabular}
    \end{sc}
    \end{small}
    \end{center}
    \vskip -0.1in
\end{table}

We evaluated our proposed DIMV method by comparing it against several state-of-the-art techniques, including MissForest \cite{stekhoven2012missforest}, GAIN \cite{yoon2018gain}, softImpute \cite{hastie2015matrix}, and ImputePCA \cite{josse2012handling}, Variational Autoencoder (VAE) \cite{collier2021vaes}. Additionally, we included four traditional but still widely used methods, namely Multiple Imputation by Chained Equation (MICE) \cite{buuren2010mice}, K-nearest neighbor imputation (KNNI), EM algorithm, and mean imputation. 

To assess the performance of the methods on randomly missing patterns, we employed four datasets from the UCI database \cite{Dua:2019}: Thyroid, Yeast, Seeds, and Iris. The missing data was randomly generated with missing rates ranging from 10\% to 80\%. The missing rate represents the proportion of deleted entries compared to the total number of entries in the input. 
For monotone missing patterns, we use two image datasets: MNIST \cite{lecun1998mnist} and Fashion-MNIST \cite{xiao2017fashion}. To simulate the missing pattern, we randomly selected 50\% of the images and deleted a section in the right corner, varying the height and width dimensions by 40\%, 50\%, and 60\%, respectively. More details about the datasets can be found in Table \ref{tab-datasets}.

In DIMV, we employed cross-validation to select the regularization term $\alpha$. We performed a grid search over $\alpha \in [0.0, 0.01, 0.1, 1.0, 10.0, 100.0]$ and set the feature selection threshold $\tau$ to its default value of 0. The initialization value, $init\_with\_zero$, was set to False. 
We assessed the methods based on their root mean squared error (RMSE) between the imputed values and the actual values in the original dataset. Additionally, we considered the running time as a crucial factor in evaluating the methods for large datasets. For DIMV, we included the cross-validation time in the overall time measurement to account for the computational cost of parameter selection.  
The experiment was conducted on an Apple M1 Pro chip with 16GB of RAM and 8 CPU cores. The source code is available via \href{https://github.com/maianhpuco/DIMVImputation}{github: https://github.com/maianhpuco/DIMVImputation}. 

\subsection{Results and Analysis}

\begin{figure*}[ht]
   \centering
     \includegraphics[width=\linewidth]{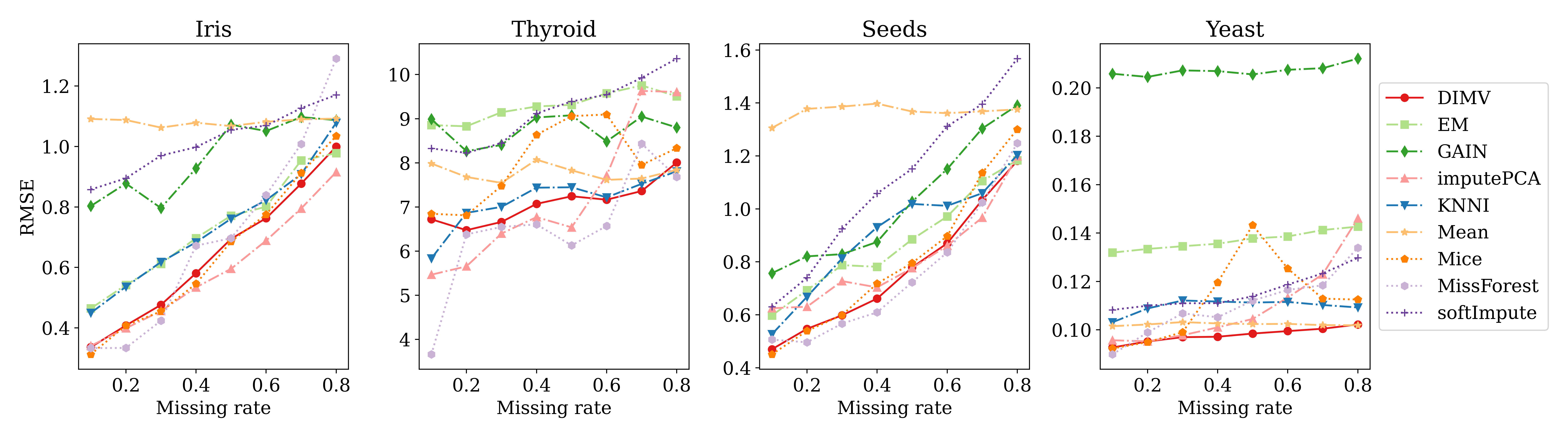}
     \caption{RMSE results on imputed tabular data}\label{Fig:Data1}
\end{figure*}

\paragraph{DIMV for randomly missing data}  Figure \ref{Fig:Data1} shows the results of tabular datasets. 
From the figure, we can see that DIMV is consistently among the top methods for all datasets. DIMV demonstrated competitive performance for the Iris dataset, closely approaching the results of ImputePCA, which has the best result for the Iris dataset. For Thyroid datasets, MissForest and ImputePCA, displayed the best performance from a missing rate of $0.1$ to $0.6$, while KNNI and DIMV yielded slightly higher RMSE at these missing rates. However, KNNI and DIMV were more stable at the high missing rate $0.7 - 0.8$. For Seeds, the top algorithms, including MissForest, DIMV, MICE, and Impute PCA, have closely aligned results. 
The small gaps in RMSE results observed between the top-performing algorithms in the figure indicate the competitiveness of ImputingPCA, MissForest, MICE, and DIMV. Based on this result, we can see that DIMV demonstrates competitive performance and robustness among widely used methods for small datasets.

\vspace{-10pt}

\paragraph{Performance under multicollinearity} Note that the $6^{th}$ and $7^{th}$ columns of the Yeast datasets are columns of constants. Therefore, we can consider this as a dataset with multicollinearity. From Figure \ref{Fig:Data1}, we see that DIMV has the best performance of all methods under comparison for missing rate $0.3-0.8$, and is also among the best ones for missing rate $0.1-0.2$.  The experiment on this dataset illustrates the capability of DIMV to handle multicollinear data compared to other techniques. Interestingly, mean imputation surpasses various state-of-the-art methods in such a multicollinear scenario.

\textbf{DIMV for monotone missing data.}  In various scenarios, the data is not missing randomly for all features but rather in the monotone pattern. For example, a specific corner of the image is missing. To investigate and gain insight into the performance of different imputation methods, we conducted an experiment using MNIST and Fashion-MNIST datasets. In this experiment, we assess the performance of the DIMV algorithm with  GAIN and VAE, EM algorithm, KNNI, softImpute, and mean imputation). 

The results are reported in Table \ref{table:mono-rmse}. KNNI shows the second-best result in the RMSE evaluation and demonstrates excellent results for the imputed plot. However, there is a massive trade-off in terms of computational time for KNNI; it took more than 2,5 hours to complete a missing rate, while other methods, including DIMV, EM, GAIN, Mean, softImpute, and VAE, only require $1/5$ th of the time. Therefore, for large datasets, KNNI is a computationally expensive option. 

Comparing methods that are suitable for a large dataset in this experiment, including DIMV, EM, GAIN, softImpute, and VAE, we can see DIMV displayed the best performance in terms of RMSE results, VAE also has a closed-aligned result in terms of computational time for high missing rate. From Table \ref{table:mono-rmse}, we can see that DIMV delivers a consistent and remarkable performance compared to the other methods, significantly when the missing rate is increased. Specifically, at 40\% missing rates for the MNIST dataset, the RMSE for DIMV  is $38.7$, while the second-best result is closely behind, which is $39.5$ from VAE; but for missing rate 60\%, the best result from DIMV is 59.8 while the second-best from VAE has a more significant gap which is 68.5. 

\begin{figure}
    \centering
    \subfigure[MNIST dataset]{
        \includegraphics[height = 6.5cm]{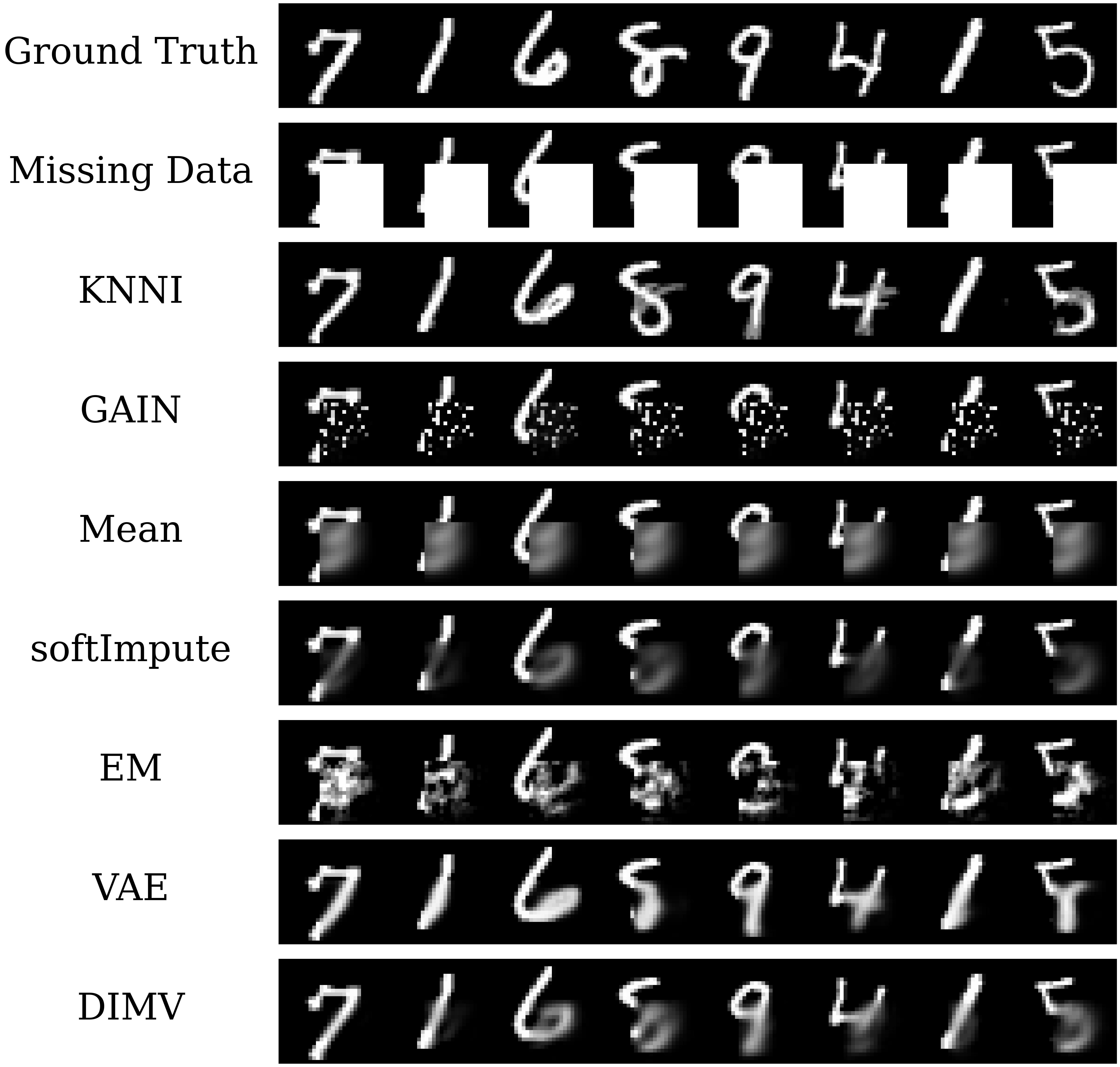}
    }
    \hspace{0 cm}
    \subfigure[Fashion-MNIST dataset]{
        \includegraphics[height = 6.5cm]{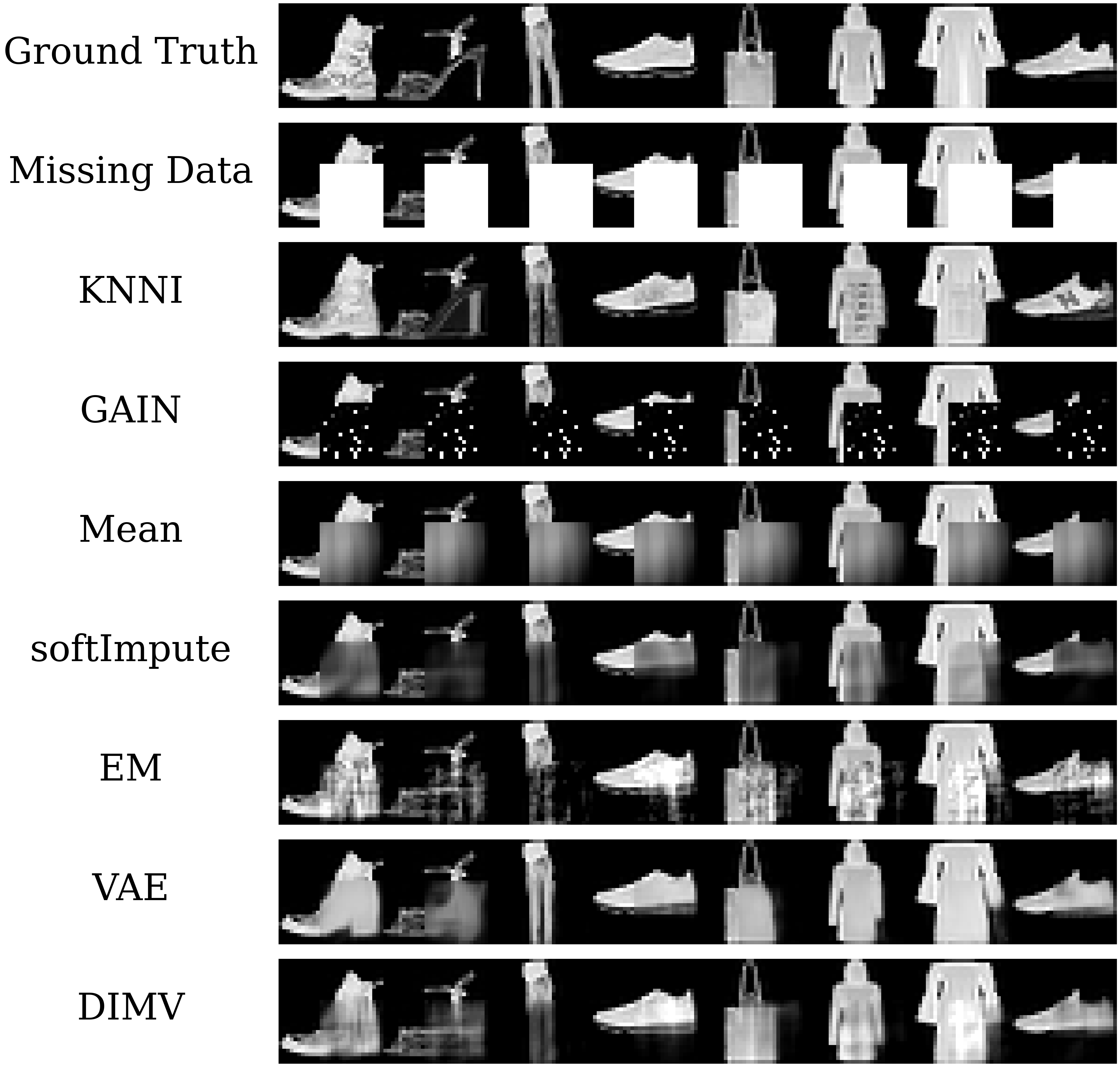} 
    }
    \caption{Plotted imputed results for monotone missing datasets}
    \label{fig:images}
    \label{fig:mono-plot}
\end{figure}

Moreover, the visualization of imputed data in Figure \ref{fig:mono-plot}, DIMV shows notably distinguishable better-imputed digits compared to the other scalable approaches under comparison.


\begin{table}[ht]
    \caption{RMSE under different missing rates. The methods that cannot deliver results in 5 hours are removed from the table.}\label{table:mono-rmse}
    \setlength{\tabcolsep}{1.3 pt}
    \vskip 0.15in
    \begin{center}
    \begin{small}
    \begin{sc}
\begin{tabular}{|c|c|c|c|c|c|c|c|c|}
\hline
      dataset & missing rate&               DIMV &    EM &   GAIN &   KNN &  mean & softImpute &   VAE \\
\hline
Fashion-MNIST & 0.4 &  $\textbf{ 34.9 }$ &   48.5 &   68.1 &  37.8 &  77.7 &       43.0 &  38.6 \\
      & 0.5 &  $\textbf{ 38.3 }$ &   53.2 &   95.8 &  40.6 &  79.6 &       48.7 &  43.0 \\
      & 0.6 &  $\textbf{ 42.1 }$ &   58.0 &  130.0 &  42.8 &  79.9 &       59.2 &  46.1 \\
\hline
MNIST & 0.4 &  $\textbf{ 38.7 }$ &   82.1 &   74.5 &  42.2 &  57.9 &       44.1 &  39.5 \\
      & 0.5 &  $\textbf{ 52.5 }$ &  100.9 &   86.4 &  55.4 &  71.2 &       60.6 &  59.8 \\
      & 0.6 &  $\textbf{ 59.8 }$ &  108.1 &  101.5 &  62.3 &  76.3 &       70.0 &  68.5 \\
\hline
\end{tabular}
    \end{sc}
    \end{small}
    \end{center}
    \vskip -0.1in
\end{table}    
 \begin{table}[ht]
    \caption{Running time (in seconds) under different missing rates. The methods that cannot deliver results in 5 hours are removed from the table. }
    \setlength{\tabcolsep}{1.3 pt}\label{table:mono-time}
    \vskip 0.15in
    \begin{center}
    \begin{small}
    \begin{sc}
\begin{tabular}{|c|c|c|c|c|c|c|c|c|}
\hline
      dataset & missing rate &    dimv &          em &    gain &      knn &              mean & softimpute &     vae \\
\hline
fashion-MNIST & 40  &  1326 & 1310 &    619 &   8281     &   2  &      398 &  1478 \\
      & 50 &  1278 &   2413 &   594 &  10870 &              2  &      893 &  1181 \\
      & 60 &  1628 &   4013   & 510 &  13137 &              2  &     1062 &  1106 \\
\hline
MNIST & 40 &   826 &   1209 &   456 &   9029 &              1  &      567 &  1665 \\
      & 50 &  1167 &   2211 &   433 &   9387 &              1  &      789 &  1071 \\
      & 60 &  1617 &  2979 &  1208 &  10540 &              2  &      921 &  1629 \\
\hline
\end{tabular}
    \end{sc}
    \end{small}
    \end{center}
    \vskip -0.1in
\end{table}    


\section{Discussion}\label{sec-discussion}
For a matrix decomposition technique such as ALS \cite{hastie2015matrix}, predicting missing entries for a sample requires stacking that sample to the training data and conducting matrix decomposition again. The proposed DIMV algorithm, however, can predict sample by sample. 
Further, if the data is centered during the missing value prediction phase, our method only requires the storage of a covariance matrix of size $p\times p$ parameters and $\alpha$. This may be tiny compared to a deep learning model for missing value imputation. Hence, it is lightweight and can be used for Internet of Things devices, mobile devices, and web browsers.

Note that DIMV relies on DPER for estimating parameters, and  DPER is for randomly missing data. In addition, the conditional Gaussian formula and multivariate Gaussian distribution is a good approximation of data distribution when the sample size is large enough. Therefore, DIMV is suitable for randomly missing data when the sample size is reasonably large compared to the number of features. If the data is skewed, however, a potential solution is to transform the data (e.g., log transform).

Regarding the threshold for DIMV, note that 
features that have a lower correlation with the features to impute can be useful. So, the purpose of using a threshold is to remove redundant features such as constant features or the features that are noises. The larger the number of features, the higher the potential for redundant features. 
Another noteworthy thing is that during the imputation process, prior knowledge of the features that affect the current feature to be imputed or the features that are collinear could also be used in choosing the features to condition upon, rather than using only the correlation as a selection criterion. 
In addition, if collinearity exists during the imputation process, then $\alpha$ should not be set to 0, as $L_2$ norm is a common tool for dealing with collinearity. 

\section{Conclusion}\label{sec-conclusion}
In this paper, we introduced DIMV, an imputation technique that finds the conditional distribution of a feature with missing entries based on fully observed features. 
As illustrated, our algorithm achieved a low RMSE compared to state-of-the-art imputation techniques, and it is robust to the assumption of multivariate normal, and does not require any label information. Hence, it can be used for both supervised and unsupervised learning. We have shown the relation of DIMV to Ridge regression and how DIMV can explain the contribution of each feature to the imputation of a feature with a missing value in a regression coefficient-like manner. In addition, the technique is robust to multicollinearity due to $L_2$ norm regularization. 

However, a limitation of DIMV is that it relies on the assumption of normal distribution and, therefore, can only be used for continuous data, and the data, if too skewed, may need to be transformed before using DIMV. Hence, in the future, we will study the effect of transformation on DIMV for skewed data. 
In addition, we are planning to extend DIMV to mixed data. Further, different parameter estimation techniques, such as DPER \cite{NGUYEN2022108082}, EPEM~\cite{NGUYEN20211}, and PMF~\cite{nguyen4260235pmf}, are worth exploring in relation to the type of data and the missing mechanism. Also, many datasets have label information readily available. Therefore, it is worth exploring a way to incorporate the label information to improve the imputation quality in a supervised setting. Moreover, when dealing with small sample sizes, exploring non-Gaussian distributions can be interesting. We also intend to investigate parallelizing DIMV to increase its speed and scalability. Lastly, the examination of DIMV's performance on datasets with collinearity, given its use of $L_2$ norm regularization, is worth exploring even more throughout, in the future. 




\bibliography{ref}

\newpage
\appendix

\section{Proof of Theorem 3.2}\label{proof-unrelated}
\textit{    Suppose that we have an observation $\mathbf{x}$ that comes from a distribution with zero mean and covariance matrix $\mathbf{\Sigma}$. Suppose that we can partition $\mathbf{x}, \mathbf{\Sigma}$ into}
    \begin{equation}
        \mathbf{x}=\begin{pmatrix}
            \mathbf{x}_o\\
            \mathbf{x}_\epsilon\\
            \mathbf{x}_m
        \end{pmatrix}, \;
        \mathbf{\Sigma}=\begin{pmatrix}
            \mathbf{\Sigma}_{o} & \mathbf{\Sigma}_{o\epsilon} & \mathbf{\Sigma}_{om}\\
            \mathbf{\Sigma}_{\epsilon o} & \sigma_\epsilon & \epsilon \\
            \mathbf{\Sigma}_{mo} & \epsilon &\sigma_m
        \end{pmatrix}        
    \end{equation}
    \textit{where $\mathbf{x}_m \in \mathbb{R}$ is a feature that contains missing values, $\mathbf{x}_o$ corresponding to the observed partition that is highly correlated with the missing entry $\mathbf{x} \in \mathbb{R}$, and $\mathbf{x}_\epsilon \in \mathbb{R}$ is observed but has a low correlation of $\epsilon$ with $\mathbf{x}_m$. In addition, $\mathbf{\Sigma}_o$ is the partition of $\mathbf{\Sigma}$ that corresponds to the observed features, $\sigma_\epsilon, \sigma_m \in \mathbb{R}$ is the variance of $\mathbf{x}_\epsilon, \mathbf{x}_m$, respectively. Also, $\mathbf{\Sigma}_{\epsilon o}, \mathbf{\Sigma}_{mo}$ is the covariance matrix between $\mathbf{x}_\epsilon$ and $\mathbf{x}_o$, $\mathbf{x}_m$ and $\mathbf{x}_o$, respectively.}

    \textit{Let $\hat{\mathbf{x}}_m = \hat{\mathbf{\Sigma}}_{om}^T \hat{\mathbf{\Sigma}}_o^{-1} \mathbf{x}_o $ be the imputed value based on conditioning on $\mathbf{x}_o$. In addition, let $\hat{\mathbf{x}}_{\epsilon}$ be the imputed value of $\mathbf{x}_m$ based on conditioning on both $\mathbf{x}_o$ and $\mathbf{x}_\epsilon$. Moreover, suppose that $\sigma_\epsilon\neq \hat{\mathbf{\Sigma}}_{\epsilon o} \hat{\mathbf{\Sigma}}_o^{-1}\hat{\mathbf{\Sigma}}_{o\epsilon}.$
    Then, the difference between the two imputed values is}
    \begin{equation}
        \hat{\mathbf{x}}_\epsilon-\hat{\mathbf{x}}_m = \frac{(\hat{\mathbf{\Sigma}}_{mo} \hat{\mathbf{\Sigma}}_o^{-1} \hat{\mathbf{\Sigma}}_{o\epsilon} -\epsilon) (\hat{\mathbf{\Sigma}}_{\epsilon o} \hat{\mathbf{\Sigma}}_o^{-1} \mathbf{x}_o-\mathbf{x}_\epsilon) }{\sigma_\epsilon -  \hat{\mathbf{\Sigma}}_{\epsilon o} \hat{\mathbf{\Sigma}}_o^{-1}\hat{\mathbf{\Sigma}}_{o\epsilon} }.
    \end{equation}
\textbf{Proof.}

Let 
\begin{equation*}
    \gamma  = \sigma_\epsilon -  \hat{\mathbf{\Sigma}}_{\epsilon o} \hat{\mathbf{\Sigma}}_o^{-1}\hat{\mathbf{\Sigma}}_{o\epsilon}. 
\end{equation*}
Then $\gamma \neq 0$ (since we assume that $\sigma_\epsilon\neq \hat{\mathbf{\Sigma}}_{\epsilon o} \hat{\mathbf{\Sigma}}_o^{-1}\hat{\mathbf{\Sigma}}_{o\epsilon}$).

Recall that $\hat{\mathbf{x}}_m = \hat{\mathbf{\Sigma}}_{om}^T \hat{\mathbf{\Sigma}}_o^{-1} \mathbf{x}_o = \hat{\mathbf{\Sigma}}_{mo} \hat{\mathbf{\Sigma}}_o^{-1} \mathbf{x}_o $ be the imputed value based on conditioning on $\mathbf{x}_o$. In addition, by using the Woodbury Matrix Identity for block matrix inversion, we have the imputed value of $\mathbf{x}_m$ based on conditioning on both $\mathbf{x}_o$ and $\mathbf{x}_\epsilon$ is
\begin{align*}
    \hat{\mathbf{x}}_{\epsilon} &= 
    ( \hat{\mathbf{\Sigma}}_{mo} \;\;\; \epsilon)
    \begin{pmatrix}
        \hat{\mathbf{\Sigma}} _o & \hat{\mathbf{\Sigma}} _{o\epsilon}\\
        \hat{\mathbf{\Sigma}}_{\epsilon o} & \sigma_\epsilon
    \end{pmatrix}^{-1}
    \begin{pmatrix}
        \mathbf{x}_o\\
        \mathbf{x}_\epsilon 
    \end{pmatrix}\\
    &= ( \hat{\mathbf{\Sigma}}_{mo} \;\;\; \epsilon) 
    \begin{pmatrix}
        \hat{\mathbf{\Sigma}}_o^{-1} + \hat{\mathbf{\Sigma}}_o^{-1} \hat{\mathbf{\Sigma}}_{o\epsilon} \hat{\mathbf{\Sigma}}_{\epsilon o} \hat{\mathbf{\Sigma}}_o^{-1}/\gamma  
        & -\hat{\mathbf{\Sigma}}_o^{-1} \hat{\mathbf{\Sigma}}_{o\epsilon}/\gamma \\
        - \hat{\mathbf{\Sigma}}_{\epsilon o} \hat{\mathbf{\Sigma}}_o^{-1}/\gamma  & 1/\gamma 
    \end{pmatrix}
    \begin{pmatrix}
        \mathbf{x}_o\\
        \mathbf{x}_\epsilon
    \end{pmatrix}\\
    &=   (\hat{\mathbf{\Sigma}}_{mo}\hat{\mathbf{\Sigma}}_o^{-1} + \hat{\mathbf{\Sigma}}_{mo}\hat{\mathbf{\Sigma}}_o^{-1} \hat{\mathbf{\Sigma}}_{o\epsilon} \hat{\mathbf{\Sigma}}_{\epsilon o} \hat{\mathbf{\Sigma}}_o^{-1}/\gamma  
    - \epsilon \hat{\mathbf{\Sigma}}_{\epsilon o} \hat{\mathbf{\Sigma}}_o^{-1}/\gamma  
     -\hat{\mathbf{\Sigma}}_{mo}\hat{\mathbf{\Sigma}}_o^{-1} \hat{\mathbf{\Sigma}}_{o\epsilon}/\gamma  + \epsilon/\gamma )
    \begin{pmatrix}
        \mathbf{x}_o\\
        \mathbf{x}_\epsilon
    \end{pmatrix}\\
    &= \hat{\mathbf{\Sigma}}_{mo}\hat{\mathbf{\Sigma}}_o^{-1} \mathbf{x}_o+ \hat{\mathbf{\Sigma}}_{mo}\hat{\mathbf{\Sigma}}_o^{-1} \hat{\mathbf{\Sigma}}_{o\epsilon} \hat{\mathbf{\Sigma}}_{\epsilon o} \hat{\mathbf{\Sigma}}_o^{-1} \mathbf{x}_o/\gamma  
    - \epsilon \hat{\mathbf{\Sigma}}_{\epsilon o} \hat{\mathbf{\Sigma}}_o^{-1} \mathbf{x}_o/\gamma  
    -\hat{\mathbf{\Sigma}}_{mo}\hat{\mathbf{\Sigma}}_o^{-1} \hat{\mathbf{\Sigma}}_{o\epsilon} \mathbf{x}_\epsilon/\gamma + \epsilon\mathbf{x}_\epsilon/\gamma \\
\end{align*}
 Therefore, 
\begin{align*}
     \hat{\mathbf{x}}_{\epsilon} -  \hat{\mathbf{x}}_m &= \frac{1}{\gamma }(\hat{\mathbf{\Sigma}}_{mo}\hat{\mathbf{\Sigma}}_o^{-1} \hat{\mathbf{\Sigma}}_{o\epsilon} \hat{\mathbf{\Sigma}}_{\epsilon o} \hat{\mathbf{\Sigma}}_o^{-1} \mathbf{x}_o 
    - \epsilon \hat{\mathbf{\Sigma}}_{\epsilon o} \hat{\mathbf{\Sigma}}_o^{-1} \mathbf{x}_o
    -\hat{\mathbf{\Sigma}}_{mo}\hat{\mathbf{\Sigma}}_o^{-1} \hat{\mathbf{\Sigma}}_{o\epsilon} \mathbf{x}_\epsilon+ \epsilon\mathbf{x}_\epsilon)\\
    &=\frac{1}{\gamma }[(\hat{\mathbf{\Sigma}}_{mo}\hat{\mathbf{\Sigma}}_o^{-1} \hat{\mathbf{\Sigma}}_{o\epsilon} -\epsilon)\hat{\mathbf{\Sigma}}_{\epsilon o} \hat{\mathbf{\Sigma}}_o^{-1} \mathbf{x}_o 
    -(\hat{\mathbf{\Sigma}}_{mo}\hat{\mathbf{\Sigma}}_o^{-1} \hat{\mathbf{\Sigma}}_{o\epsilon} \mathbf{x}_\epsilon - \epsilon )\mathbf{x}_\epsilon]\\
    &= \frac{1}{\gamma }[(\hat{\mathbf{\Sigma}}_{mo}\hat{\mathbf{\Sigma}}_o^{-1} \hat{\mathbf{\Sigma}}_{o\epsilon} -\epsilon)(\hat{\mathbf{\Sigma}}_{\epsilon o} \hat{\mathbf{\Sigma}}_o^{-1} \mathbf{x}_o -\mathbf{x}_\epsilon)]
\end{align*}

\section{An example illustrating the computation process of DIMV} \label{apd-example}
In this section, we present an example illustrating the computation process of DIMV. The covariance matrix is computed using the \textbf{DPER algorithm} on the missing data. If the train-test split is provided, the covariance matrix will be computed using the train set only. 

Suppose that we have a dataset with missing entries as follows. 
\begin{equation}
    \begin{pmatrix}
        2 &  * &  1 &  4 &  * &  * \\
        * &  * &  4 &  7 &  * &  * \\
        3 &  * &  0 &  3 &  7 &  * \\
        5 &  3 &  6 &  * &  9 &  7 \\
        * &  1 &  4 &  7 &  5 &  2 \\
    \end{pmatrix}
\end{equation}

In this dataset, each row represents a sample, and each column represents a feature. Hence, this dataset has five samples and six features. We impute this data after normalization. For the \textbf{FeatureSelection} algorithm, we have the parameters $\tau$ and set $k$ to 2. Let us consider the imputation of the feature $f$ as the $2^{nd}$ features with the samples of the $1^{st}$, $2^{nd}$, and $3^{rd}$ row using the \textbf{DIMV algorithm}.  

If $init\_with\_zeros=$ False: 
\begin{itemize}
    \item Finding the set $\mathcal{F}$: We start by looking at the $1^{st}$ value of the $2^{nd}$ feature. Let's say the $5^{th}$ and $6^{th}$ features have a correlation with this value that is greater than a predefined threshold $\tau$. However, the values in these features are missing for the considered sample, we then select the top $k=2$ features that have the highest correlation with the $2^{nd}$ feature. In this case, let's say the $3^{rd}$ and $4^{th}$ features have the highest correlation, so we form the set $\mathcal{F}$ with these features. 

    \item Stacking values to $\mathbf{Z}$: We iterate through the missing values in the $2^{nd}$ feature. For each missing value, we consider the corresponding sample (let's call it $s$) and loop through other samples with missing values in the $2^{nd}$ feature (the $2^{nd}$ and $3^{rd}$ rows). If these samples have the same missing pattern with $s$ (they have common missing values in the $5^{th}$ and $6^{th}$ features) and share commonly observed features in set $\mathcal{F}$ (the $3^{rd}$ and $4^{th}$ features), we stack these rows together. We then stack  $2^{nd}$ sample with $s$. The $3^{rd}$ sample, however, only has a missing value in the $6^{th}$ feature (not the $5^{th}$ feature); it is not stacked to $\mathbf{Z}$.
 
    \item Imputing $\mathbf{Z}$: Once we have the stacked matrix $\mathbf{Z}$, we proceed with imputing the missing values. This is done by computing the estimated values $\hat{\mathbf{Z}}_{f}$ as specified in lines 13 to 15 of the DIMV algorithm.
 
    \item Continuing the imputation: After completing the imputation for the $2^{nd}$ sample, we move on to the $3^{rd}$ sample and repeat the same process as the three steps above. 

\end{itemize}

If $init\_with\_zeros=$ True:  

\begin{itemize}
    \item  Finding the set $\mathcal{F}$: We also start by looking at the $1^{st}$ value of the $2^{nd}$ feature. All missing values are initially set to 0. We add the $5^{th}$ and $6^{th}$ features into $\mathcal{F}$ that have a correlation with this value greater than the predefined threshold $\tau$. 
    
    \item Stacking values to $\mathbf{Z}$: Next, we proceed to stack all the remaining samples in the feature $f$ into $\mathbf{Z}$.
 
    \item Imputing $\mathbf{Z}$: We compute the estimated values $\hat{\mathbf{Z}}_{f}$ as specified in lines 13 to 15 for each position in $f$. These estimated values are used as the imputed values for the missing data. 
 
\end{itemize}

\section{Accuracy results on imputed tabular data}\label{apd-acc-tabular}
\begin{figure*}[ht]
   \centering
     \includegraphics[width=.9 \linewidth]{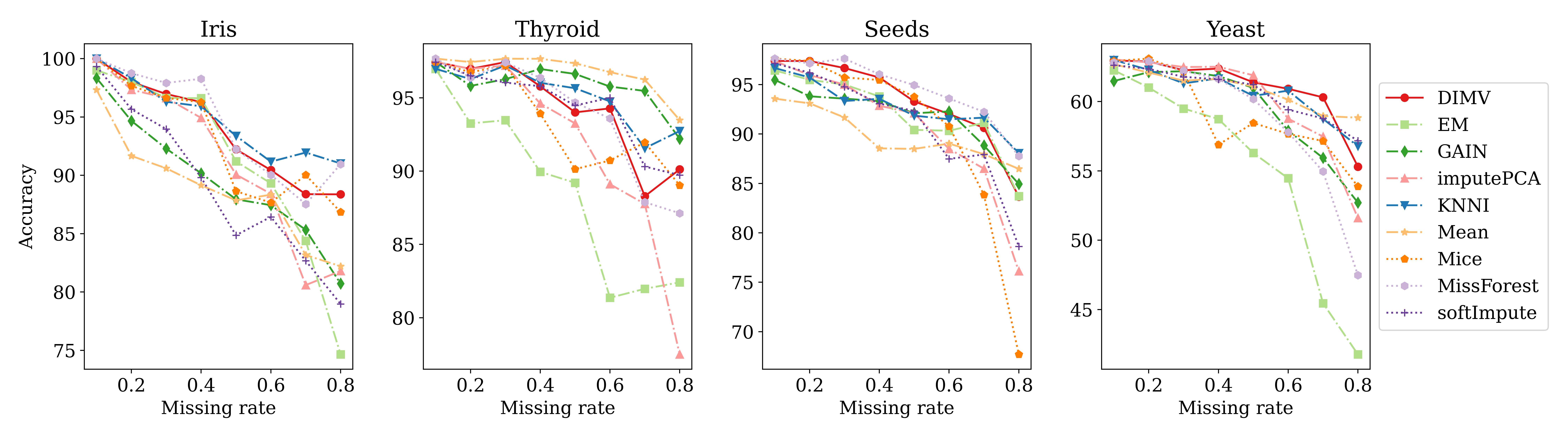}
     \caption{Accuracy on imputed tabular data.}\label{Fig:acc-random}
   \centering
\end{figure*}

The accuracy results on imputed tabular data are presented in Figure \ref{Fig:acc-random}, where DIMV is presented in red. The figure shows the competitive performance of DIMV for small datasets, especially the ability to handle multicollinear data (Thyroid datasets).

\section{Accuracy results on imputed monotone data}
The accuracy results on imputed monotone data are presented in Table \ref{tab-acc-mnist}. The results show that DIMV achieves the best performance in terms of accuracy in almost all cases.

\begin{table}[ht]
    \caption{Classification accuracies under various missing rates. The methods that cannot deliver results in 5 hours are removed from the table.}\label{tab-acc-mnist}
    \setlength{\tabcolsep}{1.3 pt}
    \vskip 0.15in
    \begin{center}
    \begin{small}
    \begin{sc}
\begin{tabular}{|c|c|c|c|c|c|c|c|c|}
\hline
      dataset & missing rate&               DIMV &    EM &   GAIN &   KNN &  mean & softImpute &   VAE \\
\hline
FashionMNIST & 40 &  84.1  &  83.9 &               83.8 &               84.0 &  84.0 &       84.0 &  73.5 \\
      & 50 &  84.1 &  84.0 &               83.8 &               84.1 &  83.9 &       84.1 &  72.5 \\
      & 60 &   84.3 &  84.0 &               83.9 &               84.2 &  83.8 &       84.0 &  72.8 \\ 
\hline

MNIST & 40 &               92.0 &  91.7 &  92.0 &               91.9 &  91.8 &               91.9 &  90.6 \\
      & 50 &   91.9 &  91.1 &               91.6 &               91.9 &  91.4 &               91.7 &  90.4 \\
      & 60 &               91.4 &  89.5 &               91.2 &  
 91.6 &  90.9 &               91.3 &  89.0 \\
\hline
\end{tabular}
    \end{sc}
    \end{small}
    \end{center}
    \vskip -0.1in
\end{table}

\section{Experiment details}
In this section, we further detail the settings for the experiments.
We use a machine with 16 GB RAM, eight cores (6 performance and 2 efficiency), an Apple M1 Pro Chip, and an SSD hard disk for all experiments. 



\textbf{Details of the settings for the techniques used in the comparison:}
\begin{itemize}
    \item \textbf{Mean}: Mean Imputation was implemented by using  \href{https://scikit-learn.org/stable/modules/generated/sklearn.impute.SimpleImputer.html}{SimpleImputer} module from scikit-learn \cite{scikit-learn} package. 
    
    \item \textbf{KNNI}: KNNI was implemented by using the \href{https://scikit-learn.org/stable/modules/generated/sklearn.impute.KNNImputer.html}{KNNImputer} module from scikit-learn \cite{scikit-learn} package with $k=5$ for randomly missing experiment and $k=2$ for monotone missing experiment. 

     \item \textbf{MICE}: MICE was implemented using the \href{https://scikit-learn.org/stable/modules/generated/sklearn.impute.IterativeImputer.html}{Iterative Imputer} package from scikit-learn\cite{scikit-learn}.

 
    \item \textbf{softImpute}: we use implementation of softImpute from package "fancyimpute" \cite{fancyimpute}.For the randomly missing experiment, we applied the 'BiScaler' preprocessing technique from the 'fancyimpute' package before applying the '.transform' function of the 'SoftImpute' method. We further used the 'inverse\_transform' function to revert the imputed data back to its original scales, as it resulted in better RMSE values, and the maximum number of iterations (max\_iter) was set to 1000. For the monotone missing experiment, we did not use the 'BiScaler' preprocessing technique, as it led to significantly higher RMSE values (greater than 10,000) for the MNIST and Fashion-MNIST datasets, the maximum number of iterations (max\_iter) was set to 100. 
    
    \item \textbf{missForest}: The missForest method was implemented using the \href{https://pypi.org/project/missingpy/}{missingpy} package in Python. The maximum number of iterations (max\_iter) was set to 100, and the criterion for imputation was set to 'squared\_error'.

    \item \textbf{GAIN}: The GAIN method was implemented based on the \href{https://github.com/jsyoon0823/GAIN}{GAIN repository} \cite{yoon2018gain}, which corresponds to the original paper. For the randomly missing data experiment, we set the parameters as follows: batch\_size: 10, hint\_rate: 0.9, alpha: 100, and iterations: 1000. For the monotone missing experiment, we used a batch size of 64, hint\_rate: 0.9, alpha: 100, and 20000 iterations.


     \item \textbf{Variational Autoencoder (VAE)}: For the VAE method, the following parameters were used for the MNIST and Fashion MNIST datasets: latent space dimension (z\_dim): 50, likelihood: 'BERNOULLI', number of mixture components: 1, and the number of epochs: 100.

     \item \textbf{EM}:  was implemented using the ``impute\_EM" function in the \href{https://www.rdocumentation.org/packages/missMethods/versions/0.2.0}{missMethods} package in R with default configuration. For randomly missing experiments, the maximum number of iterations (max\_iter) was set to 1000. The maximum number of iterations (max\_iter) was set to 10 for the monotone missing experiment.
 
     \item \textbf{ImputePCA}: was implemented using ``impute\_PCA"   from package \href{https://cran.r-project.org/web/packages/missMDA/index.html}{missMDA} in R, we use the maximum number of iterations (maxits) as 200. 
     
\end{itemize}

\textbf{Classifier}: We used a \href{https://scikit-learn.org/stable/modules/generated/sklearn.linear_model.LogisticRegression.html}{Logistics Regression} classifier with the implementation from scikit-learn \cite{scikit-learn}. The model was trained using a grid search approach with 5-fold cross-validation to determine the optimal values for the regularization parameter C, with values of [0.01, 0.1, 1, 10, 100] and penalty values in ['l1', 'l2'].
For our experimental setup, we created a train-test split for randomly missing cases and used the provided train/tests in MNIST and Fashion-MNIST datasets. We conducted the grid search process separately for each dataset on non-missing data of the train set.

After selecting the set of optimal hyperparameters, we trained logistic regression models on the imputation train set for each imputation method. We then use a test set of non-missing data to check the performance of each imputation model for the classification task.

\section{Testing multivariate normal assumption}

To check if DIMV is robust to the multivariate Gaussian assumption, we test if the datasets used follow multivariate Gaussian distribution. Specifically, we ran \emph{Royston test for Multivariate Normality} against five uncorrupted datasets to check our normality assumption. We ran the test on the entire dataset for Iris, Yeast, and Thyroid. For \emph{MNIST} \cite{lecun1998mnist} and \emph{FASHION-MNIST} \cite{xiao2017fashion}, due to the limited number of samples allowed for the test in the package function, we ran the test against 1000 samples for each dataset. None of the datasets met the multivariate normal assumption with a very stringent p-value. The result of the test is summarised in Table \ref{tab-p-value}.



\begin{table*}[ht]
    \caption{RMSE results of imputed data}
    \setlength{\tabcolsep}{1.5 pt}
    \label{tab-p-value}
    \vskip 0.15in
    \begin{center}
    \begin{small}
    \begin{sc}
\begin{tabular}{|c|c|c|}
\hline
Dataset       & Score    & p-value  \\ \hline
\hline

IRIS          & 50.37    & 3.13e-11 \\ 
YEAST         & 1241.63  & 0        \\ 
THYROID   & 315.82   & 0        \\ 
SEEDS & 64.13 & 6.04-14 \\ 
MNIST         & 96125.75 & 0        \\ 
FASHION-MNIST & 53914.19 & 0        \\  

\hline
\end{tabular}
\end{sc}
\end{small}
\end{center}
\vskip -0.1in
\end{table*}

\section{Importance of DPER for estimating parameters}

\begin{figure}[h]
\centering
    \includegraphics[scale=0.6]{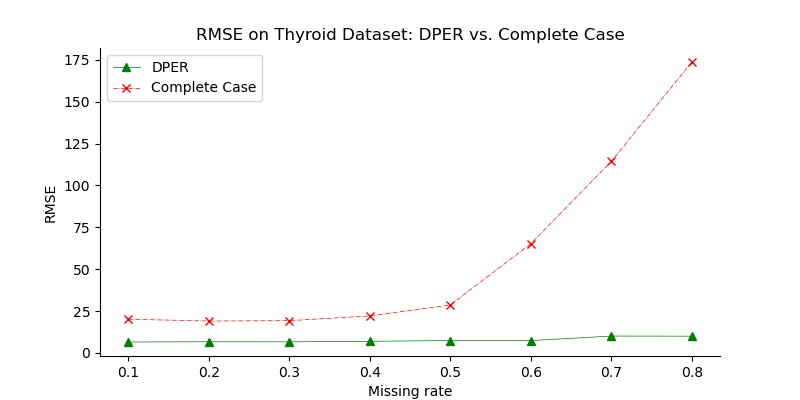}
    \caption{RMSE comparison when using Complete-case versus DPER for estimating the covariance matrix.}
\label{outliers}
\end{figure}

In this section, we experiment to compare the performance of DIMV when using DPER and complete-case analysis on the Thyroid~\cite{Dua:2019} data set. Here, we set the regularization parameter $\alpha =0$. 
The results are shown in Figure \ref{outliers}. The figure shows that the performance of DIMV, when using DPER to estimate parameters, is significantly better than using the parameters estimated from complete-case analysis. This confirms the importance of using DPER for DIMV. 

\end{document}